\documentclass[conference]{IEEEtran}
\IEEEoverridecommandlockouts
\usepackage{cite}
\usepackage{amsmath,amssymb,amsfonts}
\usepackage{graphicx}
\usepackage{textcomp}
\usepackage{xcolor}
\usepackage{caption}
\usepackage{subcaption}
\usepackage{float}
\usepackage{enumitem}
\usepackage{algorithmicx}
\usepackage[ruled]{algorithm}
\usepackage{algpseudocode}
\usepackage{footnote}
\usepackage{wrapfig}

\newcommand{\ve}[1]{\mathbf{#1}}

\def\BibTeX{{\rm B\kern-.05em{\sc i\kern-.025em b}\kern-.08em
    T\kern-.1667em\lower.7ex\hbox{E}\kern-.125emX}}

\begin{document}

\title{An Interpretable Joint Nonnegative Matrix Factorization-Based Point Cloud Distance Measure
\thanks{This work is supported by NSF DMS award \#2211318.}
}

\makeatletter
\newcommand{\linebreakand}{%
  \end{@IEEEauthorhalign}
  \hfill\mbox{}\par
  \mbox{}\hfill\begin{@IEEEauthorhalign}
}
\makeatother

\author{
\IEEEauthorblockN{Hannah Friedman}
\IEEEauthorblockA{\textit{Department of Mathematics} \\
\textit{Harvey Mudd College}\\
Claremont, CA, USA \\
hfriedman@hmc.edu}
\and

\IEEEauthorblockN{Amani R. Maina-Kilaas}
\IEEEauthorblockA{\textit{Department of Computer Science} \\
\textit{Harvey Mudd College}\\
Claremont, CA, USA \\
amainakilaas@hmc.edu}
\and
\IEEEauthorblockN{Julianna Schalkwyk}
\IEEEauthorblockA{\textit{Department of Mathematics} \\
\textit{Harvey Mudd College}\\
Claremont, CA, USA \\
jschalkwyk@hmc.edu}
\linebreakand
\IEEEauthorblockN{Hina Ahmed}
\IEEEauthorblockA{\textit{Department of Mathematics} \\
\textit{Scripps College}\\
Claremont, CA, USA \\
hahmed@hmc.edu}
\and
\IEEEauthorblockN{Jamie Haddock}
\IEEEauthorblockA{\textit{Department of Mathematics} \\
\textit{Harvey Mudd College}\\
Claremont, CA, USA \\
jhaddock@hmc.edu}
}

\maketitle

\begin{abstract}
In this paper, we propose a new method for determining shared features of and measuring the distance between data sets or point clouds. 
Our approach uses the joint factorization of two data matrices $X_1,X_2$ into non-negative matrices $X_1 = AS_1, X_2 = AS_2$ to derive a similarity measure that determines how well the shared basis $A$ approximates $X_1, X_2$.
We also propose a point cloud distance measure built upon this method and the learned factorization.
Our method reveals structural differences in both image and text data.
Potential applications include classification, detecting plagiarism or other manipulation, data denoising, and transfer learning.
\end{abstract}

\begin{IEEEkeywords}
nonnegative matrix factorization, topic modeling, point cloud distance, data set distance
\end{IEEEkeywords}


\section{Introduction}
Identifying similar or dissimilar features in the underlying structures of point clouds is a useful technique across a wide array of fields. 
%
Point cloud comparison methods and point cloud distances have applications in document clustering~\cite{mccrae2018linking,shahnaz2006document} and in computer vision such as object classification~\cite{qi2017pointnet}, object detection~\cite{qi2019deep}, and semantic segmentation~\cite{pham2019jsis3d}.
These applications often turn to classical metrics for dataset or point cloud comparison, such as the Chamfer distance~\cite{wu2021balanced}, which is defined as the sum of the averages of the minimum distances between points in data matrices $X_1$ and $X_2$,
\begin{align}
d_{\text{cham}}(X_1, X_2) = \frac{1}{|X_1|} &\sum_{\ve{x} \in X_1} \min_{\ve{y} \in X_2} \|\ve{y} - \ve{x}\|_2^2 \label{eq:cham}
\\ &+ \frac{1}{|X_2|} \sum_{\ve{y} \in X_2} \min_{\ve{x} \in X_1} \|\ve{x} - \ve{y}\|_2^2,\nonumber
\end{align}
where, in a slight abuse of notation, we let $\ve{x} \in X$ mean that $\ve{x}$ is a column of the data matrix $X$.
Other popular metrics include the
Hausdorff distance~\cite{huttenlocher1993comparing, javaheri2020generalized}, the earth mover's distance~\cite{mallows1972note}, and recent metrics based on optimal transport~\cite{alvarez2020geometric}
However, 
these measures can be computationally expensive or ineffective~\cite{nguyen2021point}.
Recent efforts for point cloud comparison include density-aware approaches~\cite{urbach2020dpdist, wu2021density}.  Additionally, data set or point cloud distances have applications in measuring generalizability of machine learning models~\cite{khambete,BenDavid2006AnalysisOR,mansour_2009} and transfer learning~\cite{liu2022wasserstein}.


Nonnegative matrix factorization (NMF) is a useful tool for interpretable dimension reduction. Many types of data, including documents and images, can be represented by nonnegative matrices, making NMF a widely applicable method for data analysis \cite{LeeandSeung}. 
NMF has been previously used in the creation of data set similarity metrics: Shahnaz et al.~\cite{shahnaz2006document} clusters semantic features or topics in document data and uses NMF to preserve nonnegativity. Liu et al.~\cite{liu2013multi} introduces a multi-view clustering approach based on NMF.

Joint non-negative matrix factorization (jNMF) allows for joint factorization of two data sets with a common basis \cite{kim2015simultaneous, ma2018community}. Additionally, jNMF for topic modeling has been used to identify similarities across data sets. Kim et al.~\cite{kim2015simultaneous} proposes a jNMF method for identifying both common and distinct topics among document data sets.

In this paper, we propose a new method for evaluating similarity between a pair of distinct point clouds or data sets.
Our method analyzes the outputs of jNMF and measures the contributions of the basis vectors to each set.
In our method, we first run jNMF on the two data matrices and then, motivated by statistical distribution comparison tests, we compare the empirical distribution functions that represent the jNMF coefficient factor matrices. We use this learned information to propose a point cloud distance measure.  In addition to a point cloud distance measure, our method provides information about how the data are structurally similar and dissimilar via the learned jNMF basis vectors and their measured association with each data set.

In Section II, we give a brief overview of NMF and jNMF.
In Section III, we propose a method for determining shared features in data sets, a distance measure based on this method, and we list some desirable properties of a distance measure.
In Section IV, we present examples of this method applied to real world data and experimentally verify our desired properties.

\section{Overview of NMF and jNMF}
Given a nonnegative $m \times n$ matrix $X$, the goal of non-negative matrix factorization (NMF) is to find nonnegative matrices $A$ and $S$ such that
$$X \approx AS$$
where $A$ is $m \times k$ and $S$ is $k \times n$~\cite{LeeandSeung}.
One typically chooses $k$ so that $AS$ is a low rank approximation of $X$; there are many heuristics 
for choosing the model rank $k$, which are beyond the scope of this paper.
NMF produces $A$ and $S$ by attempting to minimize the non-convex objective function,
$$\|X - AS\|_F^2 = \sum_{ij}(X_{ij} - (AS)_{ij})^2.$$
NMF models can be trained with many methods.
One of the most popular methods is multiplicative updates, which is a variant of gradient descent that ensures entrywise non-negativity in the factors \cite{algsforNMF}.
We typically interpret columns of $A$ as a set of ``basis" vectors and the $i$th column of $S$ as the coefficients of the conic combination of those basis vectors that approximates the $i$th data point.
We do not focus on how basis vectors are combined to create individual elements, but instead analyze the rows of $S$ to measure the contribution of each basis vector to the entire data set.

Joint NMF (jNMF) finds low-rank, non-negative approximations for two matrices, $X$ and $Y$, that share a common factor matrix~\cite{kim2015simultaneous}. When applied to supervised NMF, jNMF typically factors $X$ (the data) and $Y$ (e.g., class labels) as $X \approx A_1 S$ and $Y \approx A_2 S$, so that $S$ is shared between the  factorizations.
To control the emphasis put on the labels, a weighting factor $\lambda$ can be introduced into the objective function, $$\|X - A_1S\|_F^2 + \lambda \|Y-A_2S\|_F^2,$$ but we focus on the cases where the approximation terms are weighted equally, $\lambda = 1$.
When $\lambda = 1$, the factorization can easily be learned by performing NMF on the matrix obtained by stacking $X$ on top of $Y$, resulting in the factorization $$\begin{bmatrix} X \\ Y \end{bmatrix} = \begin{bmatrix} A_1\\A_2 \end{bmatrix}S.$$
Lee et al.~\cite{lee2009semi} and Haddock et al.~\cite{haddock} apply jNMF to semi-supervised tasks like document classification.

\begin{figure}
    \centering
    \includegraphics[width=0.35\textwidth]{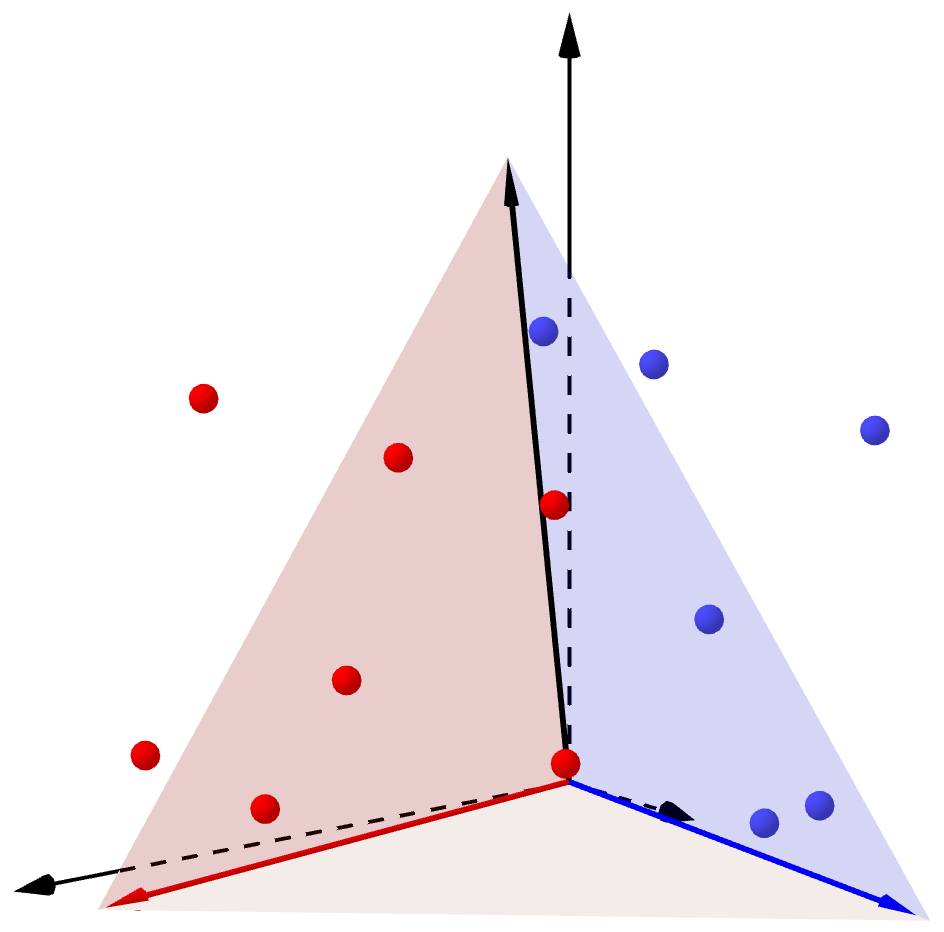}
    \caption{Visualization of a joint NMF learned for two datasets ($X_1$ in red and $X_2$ in blue).  Note that the data points in $X_1$ are well approximated by the basis elements visualized in black and red (their conic span is given in red), while the data points in $X_2$ are well approximated by the basis elements visualized in black and blue (their conic span is given in blue).}
    \label{fig:jNMF_vis}
\end{figure}

We apply jNMF via NMF on the matrix obtained by stacking data matrices $X_1$ and $X_2$ side-by-side, denoted by $[X_1\ X_2]$, which we factorize by $[X_1\ X_2] = A[S_1\ S_2]$. This model may be represented as minimizing objective
\begin{equation}\|X_1 - AS_1\|_F^2 + \|X_2 - AS_2\|_F^2\label{eq:jNMF}\end{equation}
with respect to the factor matrices $A, S_1, S_2$.
For the method to be well-defined, the data points in the two sets $X_1,X_2$ must have the same dimension.  Geometrically, NMF can be interpreted as learning basis vectors such that the cone of these vectors best approximates a given data set~\cite{gillis2020nonnegative}.  Thus, jNMF attempts to learn basis vectors (columns of $A$) so that the cone of these vectors contains good approximations of all data points in $X_1$ and $X_2$; see Figure~\ref{fig:jNMF_vis} for a visualization of a joint NMF learned for two datasets ($X_1$ in red and $X_2$ in blue).

\section{Proposed Method and Distance Measure}

Our proposed distance measure identifies how well two data sets, $X_1 \in \mathbb{R}^{m \times n_1}_{\ge 0}$ and $X_2 \in \mathbb{R}^{m \times n_2}_{\ge 0}$, can be approximated by the common basis learned through jNMF. The existence of such a set of basis vectors implies a similar underlying structure between the data sets. However, one may obtain a basis set in which some elements primarily contribute to the first data set and some contribute to the second, but very few are shared; see Figure~\ref{fig:jNMF_vis} for a visualization of such a scenario. This scenario suggests some structural differences in the data. We analyze the contributions of the basis vectors to the different data sets to identify features in one data set that are not expressed well by a basis for the other data set.

\subsection{Proposed Similarity Method}\label{subsec:method}

Given a rank-$k$ jNMF approximation $[X_1\ X_2] \approx A[S_1\ S_2]$, our method produces a length-$k$ vector $\bar{\ve{p}}$ where each element represents the ratio of the corresponding basis vector's contribution to each of the data sets.
We compute $\bar{\ve{p}}_i$ from the $i$th rows of $S_1$ and $S_2$, because
the magnitudes of the entries in row $i$ of $S_1$ and $S_2$ indicate how much $A_{:,i}$, the $i$th basis vector, contributes to each data matrix.
The entries of $\bar{\ve{p}}$ are between -1 and 1; $\bar{\ve{p}}_i$ is positive if $A_{:,i}$ contributes more to $X_1$ and negative if $A_{:,i}$ contributes more to $X_2$.  Pseudocode for our method is provided in Algorithm~\ref{alg:method}.

\begin{savenotes}
\begin{algorithm}[H]
	\caption{jNMF similarity}
	\label{alg:method}
	\begin{algorithmic}[1]
		\Require data matrices $X_1 \in \mathbb{R}^{m \times n_1}_{\ge 0}$ and $X_2 \in \mathbb{R}^{m \times n_2}_{\ge 0}$, number of samples $K$ for averaging, model rank $k$
            \State Scale each column in $X_1, X_2$ to be mean one.\label{method_step_1}\footnote{To address the case where $\ve{v} \approx \ve{0}$, we add a threshold so that $\ve{v} \in X_i$ is only normalized if $||\ve{v}|| \geq 0.05*\text{avg}_{\ve{u} \in X_i} ||\ve{u}||$.}
            \State Learn rank-$k$ jNMF approximation via~\eqref{eq:jNMF}, $$[X_1 \ X_2] \approx A[S_1\ S_2].$$
            \State For $i = 1, \cdots, k$, define $$s_i = \max\left(\{s_{ij}^{(1)}\}_{j=1}^{n_1} \cup \{s_{ij}^{(2)}\}_{j=1}^{n_2}\right)$$ where $s_{i1}^{(1)}, s_{i2}^{(1)}, \cdots, s_{in_1}^{(1)}$ and $s_{i1}^{(2)}, s_{i2}^{(2)}, \cdots, s_{in_2}^{(2)}$ are the entries of the $i$th rows of $S_1$ and $S_2$, respectively.
		\For{$j = 1, ..., K$}
            \For{$i = 1, \cdots, k$}
		\State Choose $T_i \sim \text{unif}([0,s_i]) \text{ for } i = 1, 2, \cdots, m.$\label{method_step_3}
		\State Compute $\ve{p}_i^{(j)} := F_{i}^{(2)}(T_i)- F_{i}^{(1)}(T_i)$, where $$F_i^{(1)}(T_i) := \frac{1}{n_1} \sum_{j=1}^{n_1} \mathbf{1}[s_{ij}^{(1)} < T_i]$$ and  $$F_i^{(2)}(T_i) := \frac{1}{n_2} \sum_{j=1}^{n_2} \mathbf{1}[s_{ij}^{(2)} < T_i]$$ are the empirical distribution functions (EDFs) of $\{s_{ij}^{(1)}\}_{j=1}^{n_1}$ and $\{s_{ij}^{(2)}\}_{j=1}^{n_2}$ evaluated at $T_i$, respectively.\label{method_step_4}
		\EndFor
            \EndFor
            \State \Return $\bar{\ve{p}} = \frac{1}{K} \sum_{j=1}^K \ve{p}^{(j)}$
	\end{algorithmic}
\end{algorithm}
\end{savenotes}


We note that Step~\ref{method_step_4} in the Algorithm~\ref{alg:method} compares the fraction of sample $\{s_{ij}^{(1)}\}_{j=1}^{n_1}$ below a randomly sampled threshold to the fraction of sample $\{s_{ij}^{(2)}\}_{j=1}^{n_2}$ below the same threshold.  That is, we measure the difference between these samples by calculating the difference between their EDFs $F_{i}^{(1)}(T)$ and $F_{i}^{(2)}(T)$~\cite{dekking2005modern}; this is akin to the fundamental idea of the Kolmogorov-Smirnov test~\cite{kolmogorov1933sulla,smirnov1948table} and Cramer-von Mises criterion~\cite{cramer1928composition,von1928statistik}.  See Figure~\ref{fig:hist_and_edf} for an example visualization of the samples $\{s_{ij}^{(1)}\}_{j=1}^{n_1}$ and $\{s_{ij}^{(2)}\}_{j=1}^{n_2}$, and their empirical distribution functions $F_{i}^{(1)}(T)$ and $F_{i}^{(2)}(T)$; these histograms and EDFs might correspond to the third (blue) basis vector in Figure~\ref{fig:jNMF_vis}, as the entries indicate that this basis vector is more heavily used by the blue data set. Note that $\bar{\ve{p}}_i$ is simply the difference between these EDFs averaged over random samples taken uniformly from the data interval.

\begin{figure}[h]
    \centering
    \includegraphics[width=0.5\columnwidth]{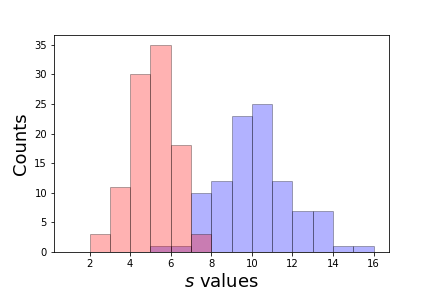}\includegraphics[width=0.5\columnwidth]{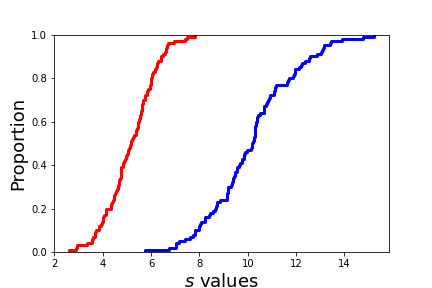}
    \caption{Example histogram of the values $s_{i1}^{(1)}, \cdots, s_{in_1}^{(1)}$ and $s_{i1}^{(2)}, \cdots, s_{in_2}^{(2)}$ encountered in Step~\ref{method_step_3} and the corresponding empirical distribution functions $F_{i}^{(1)}(T)$ and $F_{i}^{(2)}(T)$.}
    \label{fig:hist_and_edf}
\end{figure}

We additionally note that while in Step~\ref{method_step_3} of Algorithm~\ref{alg:method} we choose to uniformly sample from the interval $[0,s_i]$, one could instead evenly subdivide this interval and iterate through these break points or instead iterate through the ordered values $\{s_{ij}^{(1)}\}_{j=1}^{n_1}$ and $\{s_{ij}^{(2)}\}_{j=1}^{n_2}$.

\textbf{Example.} For illustrative purposes, suppose a rank 3 jNMF approximation of datasets $X_1$ and $X_2$ produces the vector $\bar{\ve{p}} = [-0.500, 0.001, 0.998]$.
The first basis vector contributes to both data sets, but appears with higher frequency in $X_2$, the second basis vector contributes equally to both data sets, and the third basis vector appears almost exclusively in $X_1$.  In the toy visualization of Figure~\ref{fig:jNMF_vis}, the entries of $\bar{\ve{p}} = [-0.500, 0.001, 0.998]$ might correspond to the blue, black, and red basis vectors, respectively.

\subsection{Distance Measure}\label{subsec:distance}
Although the basis vectors and the $\bar{\ve{p}}$ vector are interpretable, a single scalar value that measures similarity or distance between two data sets is often useful.
We define the distance measure between the two data matrices $X_1, X_2$ as
\begin{align*}
    d(X_1, X_2) := \|\bar{\ve{p}}\|_1,
\end{align*}
where $\bar{\ve{p}}$ is computed via the Algorithm~\ref{alg:method} in Subsection \ref{subsec:method}.

We list here a few desirable properties of a distance measure $d(X_1, X_2)$, which are satisfied by the Chamfer distance, and of the vector $\bar{\ve{p}}$.  These properties will be experimentally verified for our proposed measure in Subsection~\ref{subsec:properties}.
Let $X_1$ be a data matrix with $n$ columns.
\begin{enumerate}[label=(\textbf{P\arabic*})]
    \item\label{itm:symmetry} Symmetry: $d(X_1, X_2) = d(X_2, X_1)$ with $\bar{\ve{p}}_1 = -\bar{\ve{p}}_2$ where $\bar{\ve{p}}_1$ corresponds to the comparison $d(X_1, X_2)$ and $\bar{\ve{p}}_2$ corresponds to the comparison $d(X_2, X_1)$.
    \item\label{itm:self-similarity} Self-similarity: $d(X_1, X_1) = 0$ and $\bar{\ve{p}} = \ve{0}$.
    \item\label{permutation-invariance} Permutation invariance: If $P_\pi$ is an $n \times n$ permutation matrix,
    $d(X_1, X_1P_\pi) = 0$ and $\bar{\ve{p}} = \ve{0}$.
    \item\label{scaling-invariance} Scaling invariance: $d(X_1, \lambda X_1) = 0$, $\bar{\ve{p}} = \ve{0}$ for $\lambda > 0$.
    \item\label{large-subsets} Large subsets: If the columns of
    $\Tilde{X}_1$ are a large subset of those of $X_1$, then $d(X_1, \Tilde{X}_1) \approx 0$, and $d(X_1, \Tilde{X}_1)$ decreases monotonically as the number of columns of $\Tilde{X}_1$ approaches $n$.
    \item\label{additive-noise} Additive noise: If $\epsilon > 0$ is small and $N$ is a noise matrix, $d(X_1,  X_1 + \epsilon N) \approx 0$ and $d(X_1, X_1 + \epsilon N)$ grows monotonically with $\epsilon$.
\end{enumerate}

\section{Empirical Results}

In this section, we illustrate the proposed method and distance measure on a toy image dataset called the Swimmer dataset~\cite{donoho2003does}, which is composed of $11 \times 20$-pixel images such as that of Figure~\ref{swimmer}, and the 20 Newsgroups dataset~\cite{20news}.
Let $X_1$ be the matrix with columns that are the vectorized images from the Swimmer dataset and let $N$ be a noise matrix of the same size as $X_1$, with entries sampled i.i.d. from unif([0,1]).
All Swimmer jNMF experiments are run with rank $k = 10$.


\subsection{Distance measure properties}\label{subsec:properties}
In this section, we verify some of the desired properties from Subsection~\ref{subsec:distance} experimentally.
We note that our distance measure $d(X_1, X_2)$ appears to exhibit the self-similarity property \ref{itm:self-similarity}, the permutation invariance property \ref{permutation-invariance}, and the scaling invariance property \ref{scaling-invariance}.
Our measure, like the Chamfer measure, produces $d(X_1, X_2) = 0$ and $\bar{\ve{p}} = \ve{0}$ when applied to $X_1$ and $X_2$, an identical, permuted, or scaled copy of $X_1$. Note that we scale the data prior to applying either distance measure as indicated in Step~\ref{method_step_1} of Algorithm~\ref{alg:method}.  See Table \ref{table}.

\begin{table}[htbp]
    \centering
    \caption{Average value of our distance measure, $d(X_1,X_2)$, and the Chamfer distance, $d_{\textnormal{cham}}(X_1, X_2)$, over fifty trials, where $P_\pi$ represents the permutation matrix corresponding to a randomly sampled permutation $\pi$, and $\lambda > 0$ represents a randomly sampled value in $\{0.1, 1, 10, 100\}$, $\Tilde{X}_1$ is $X_1$ with 10\% of its columns randomly removed, and $N$ is a matrix with entries i.i.d. from unif([0, 1]).}
    \begin{tabular}{c|c|c|c|c|c|c}
         $X_2$ & $X_1$ & $X_1P_\pi $ & $\lambda X_1$ & $\Tilde{X_1}$ & $X_1 + N$ & $N$\\
         \hline

       $d(X_1,X_2)$  & 0.000 & 0.000 & 0.000 & 0.052 & 1.509 & 2.297\\
      $d_{\textnormal{cham}}(X_1, X_2)$ & 0.000 & 0.000 & 0.000 & 0.000 & 0.741 & 1.560
    \end{tabular}
    \label{table}
\end{table}

It appears that the distance measure exhibits the large subsets property \ref{large-subsets} since $d(X_1, \Tilde{X}_1)$ remains small when $\Tilde{X}_1$ is formed as a large column subset of $X_1$.
While $d(X_1, \Tilde{X}_1)$ is small, our method is still able to distinguish between $X_1$ and $\Tilde{X}_1$, whereas $d_{\textnormal{cham}}(X_1, \Tilde{X}_1) = 0$ for all the $\Tilde{X}_1$ we examine.
To verify this, we form $\Tilde{X}_1$ as a random sample of $q\%$ of columns in $X_1$, where $q \in [88,98]$ and plot $d(X_1,X_2)$ and $d_{\textnormal{cham}}(X_1, \Tilde{X}_1)$ for each value of $q$; see Figure~\ref{largesubset}.
The distance measure also appears to exhibit the additive noise property \ref{additive-noise}.
In Figure~\ref{monotone_random}, we see that $d(X_1, X_1 + \epsilon N)$, like $d_{\textnormal{cham}}(X_1, X_1 + \epsilon N)$, grows monotonically with 
$\epsilon \in [0,1]$.
All experimental values are averaged over fifty trials.

\begin{figure}[htbp]
    \centering
    \includegraphics[width = 0.4\textwidth]{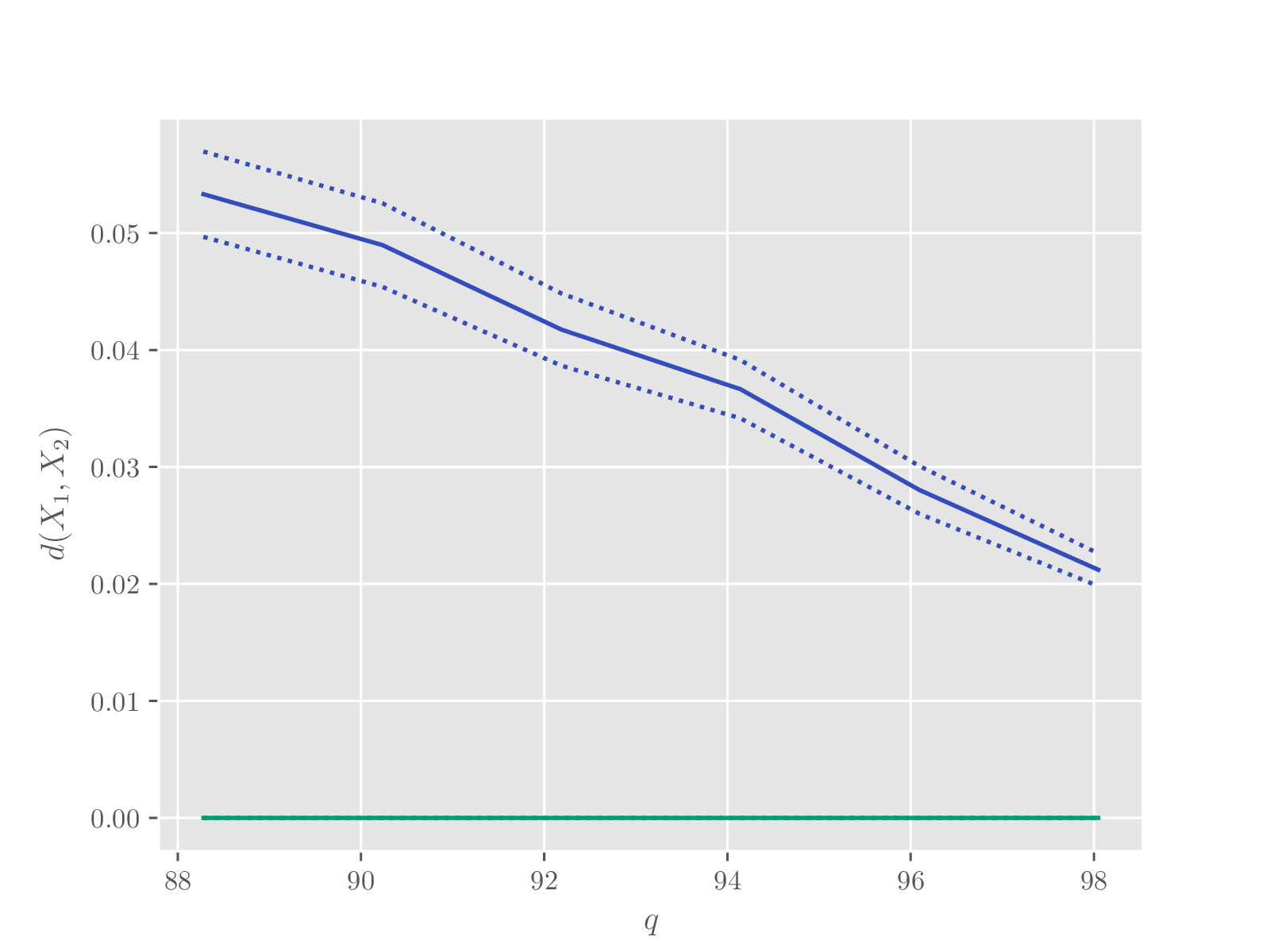}
    \caption{The jNMF distance measure (blue) and Chamfer distance (green) $d(X_1, \Tilde{X}_1)$ where $X_1$ is the Swimmer data matrix and $\Tilde{X}_1$ is formed as a random sample of $q\%$ of columns in $X_1$, for $q \in [88, 98],$ decrease monotonically as the $q$ grows. Values are averaged over fifty trials.}
    \label{largesubset}
    \vspace{-0.5cm}
\end{figure}

\begin{figure}[htbp]
    \centering
    \includegraphics[width = 0.4\textwidth]{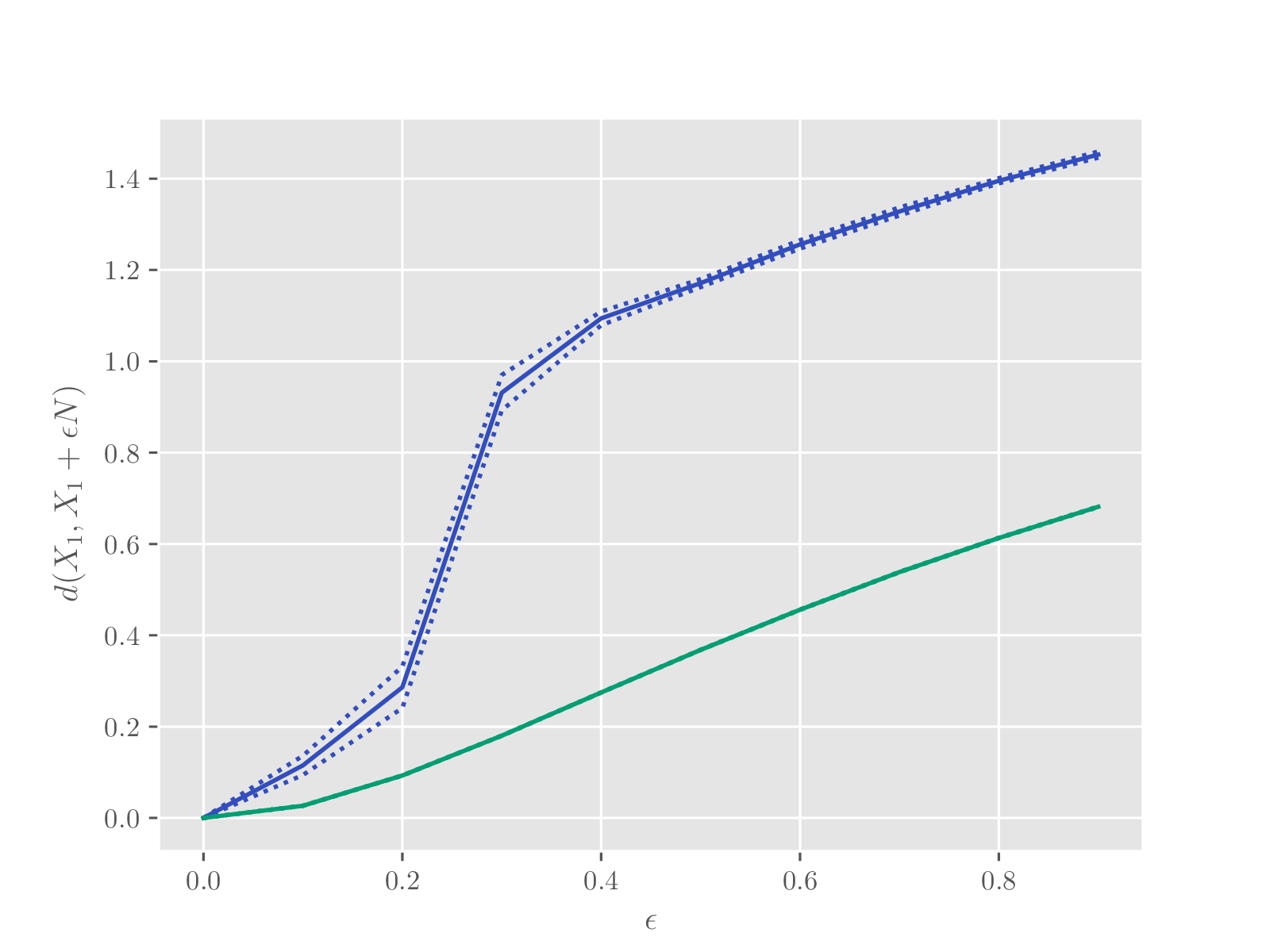}
    \caption{The jNMF distance measure (blue) and Chamfer distance (green) $d(X_1, X_1 + \epsilon N)$, where $X_1$ is the Swimmer data matrix, $N$ is a matrix with entries sampled i.i.d.\ from $\text{unif}([0, 1])$, and $\epsilon \in [0, 1]$, grow monotonically with $\epsilon$. Values are averaged over fifty trials.
}
    \label{monotone_random}
\end{figure}
Our method not only satisfies convenient distance properties like the Chamfer method, but also provides additional insight into the relationships between the datasets via the basis vectors produced by jNMF.
Figure~\ref{Xwithnoise} shows the basis vectors and their associated $\bar{\ve{p}}$ scores produced by our method applied to $X_1$ and $X_1 + N$.
The images with the dark background are good approximations for the basis vectors of the Swimmer data set.
The primarily white basis vector is used almost exclusively in the noisy data, so this vector is the primary difference between our two data sets.
The method was able to isolate the basis vectors for the original dataset from the noise.

\begin{figure}[htp]
    \centering
    \includegraphics[scale=0.1, angle=90]{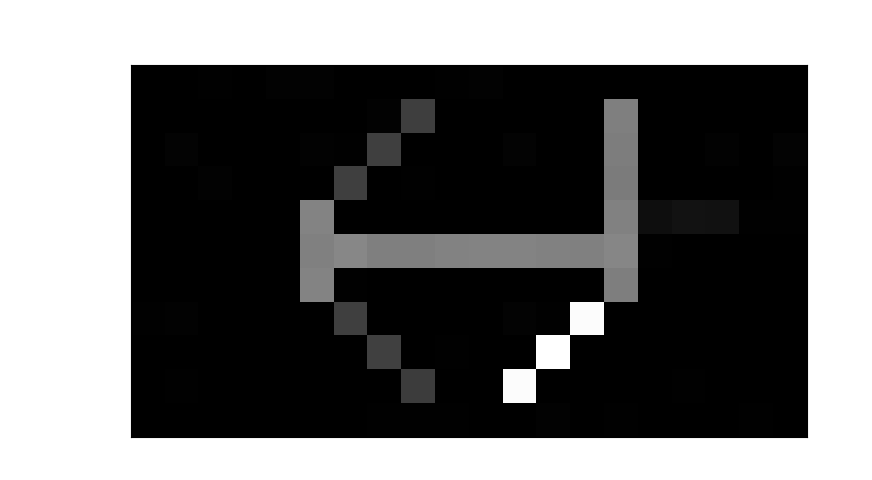}
    \includegraphics[scale=0.1, angle=90]{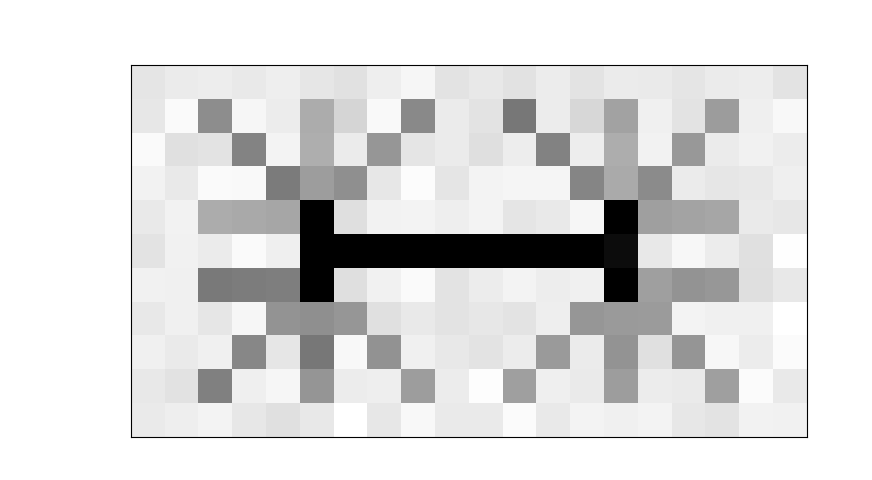}
    \includegraphics[scale=0.1, angle=90]{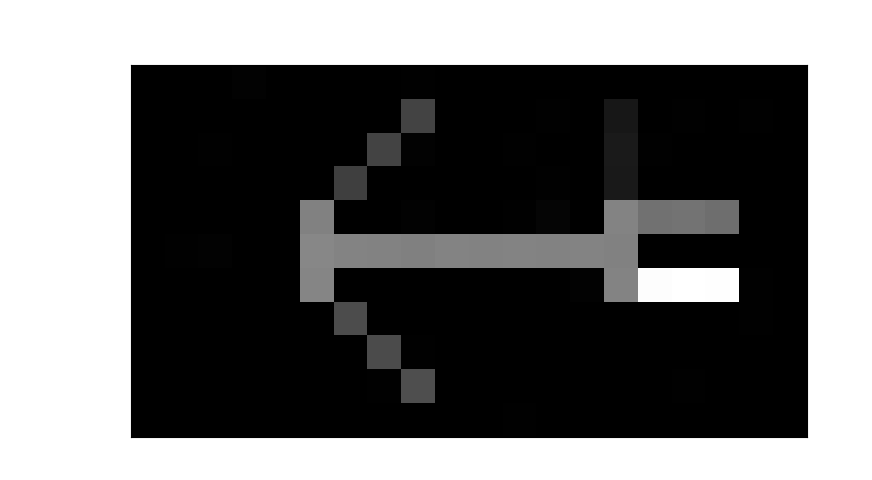}
    \includegraphics[scale=0.1, angle=90]{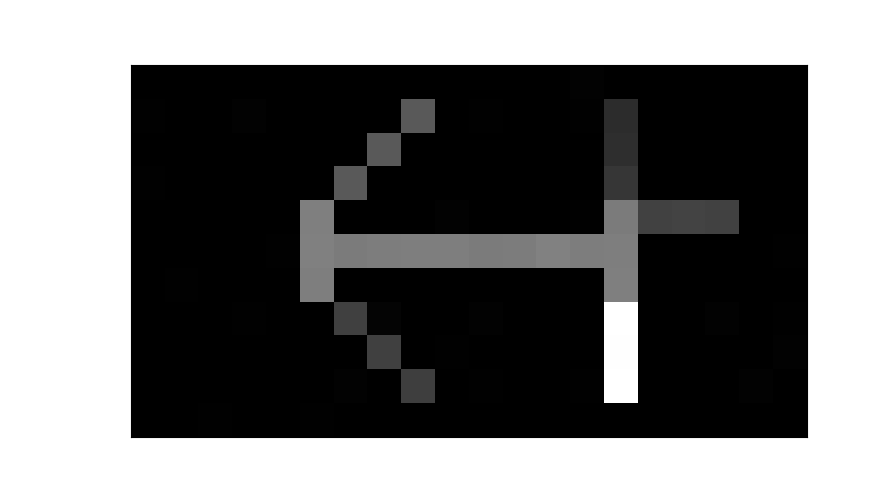}
    \includegraphics[scale=0.1, angle=90]{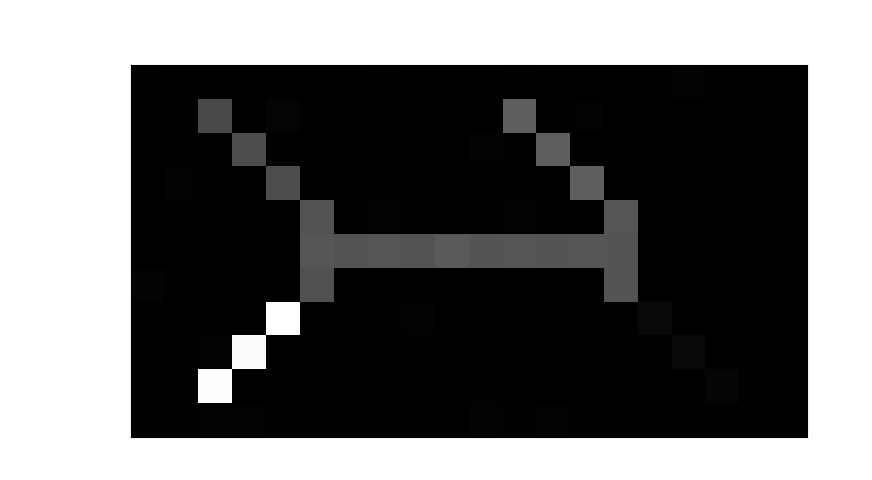}\\
    \includegraphics[scale=0.1, angle=90]{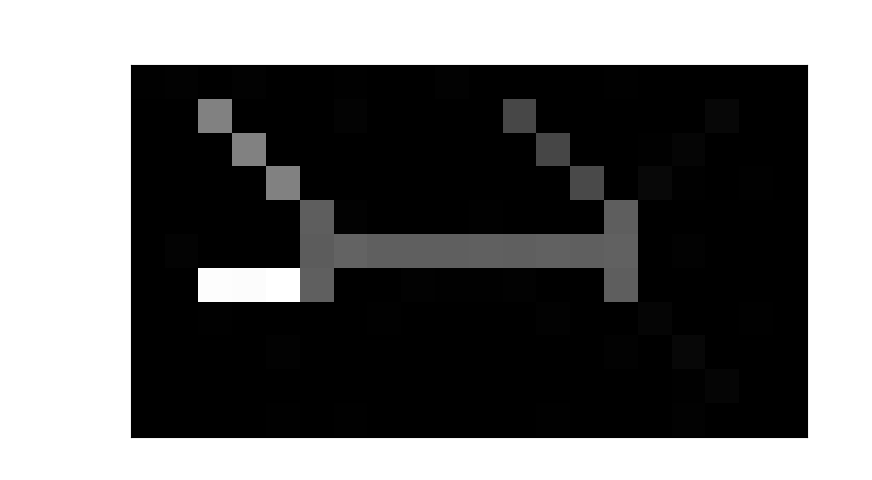}
    \includegraphics[scale=0.1, angle=90]{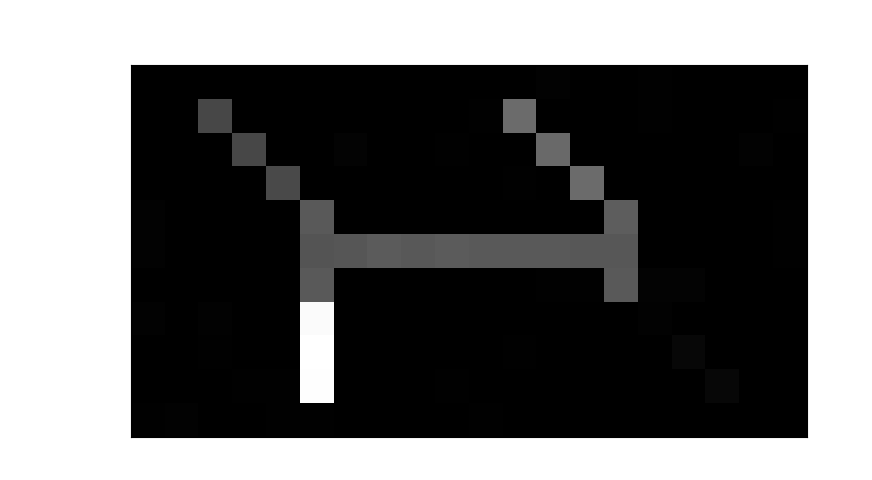}
    \includegraphics[scale=0.1, angle=90]{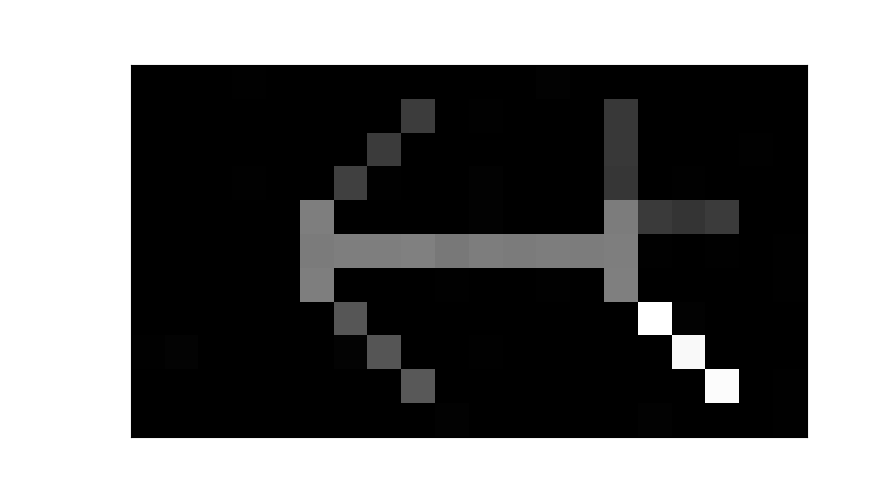}
    \includegraphics[scale=0.1, angle=90]{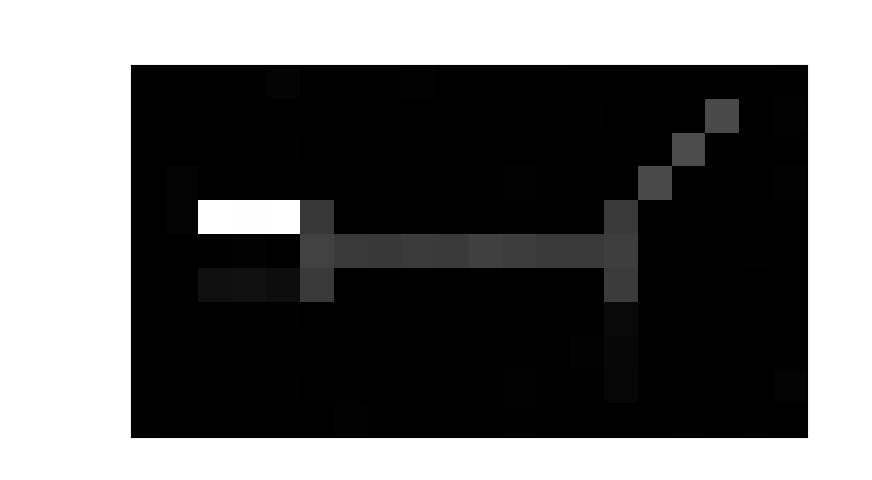}
    \includegraphics[scale=0.1, angle=90]{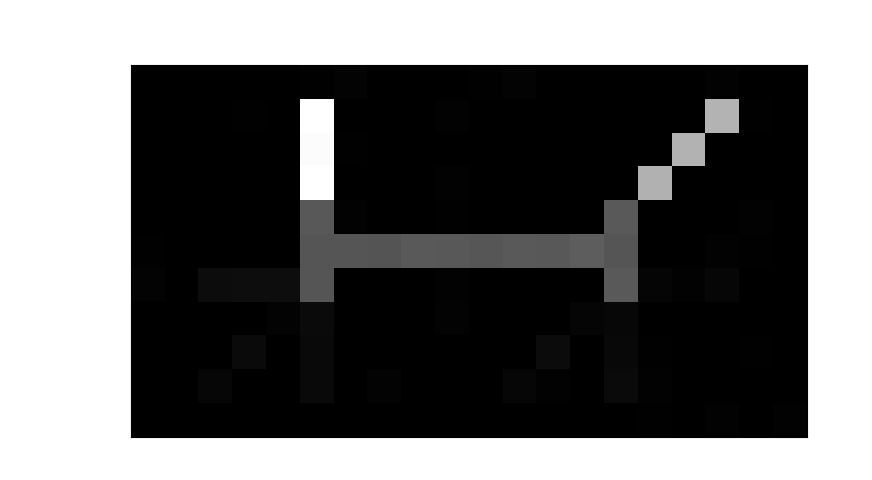}
    \caption{Basis vectors produced by our  method applied to Swimmer data matrix $X_1$ and $X_1 + N$ where the entries in $N$ are sampled i.i.d.\ from $\text{unif}([0,1])$. The associated $\bar{\ve{p}}$ values for each basis image are, reading left to right and top to bottom, $[ 0.063, -0.901,  0.076,  0.065,  0.069,  0.058,  0.058,  0.069, $ $ 0.079,  0.079]$ and $d(X_1, X_1 + N) = 1.517$ (note that this value is computed on a single trial while the corresponding entry of Table~\ref{table} is averaged over 50 trials). All basis vectors contribute roughly  equally to both data sets, with the exception of the basis vector in the second position of the first row, which contributes almost exclusively to the noisy data set $X_1 + N$.}
    \label{Xwithnoise}
\end{figure}

\subsection{Swimmer experiments}\label{subsec:swimmer}
Before we can use our measure to compare data sets, we must consider what it means for $d(X_1, X_2)$ to be small.
Since $0 \leq |\bar{\ve{p}}_i| \leq 1$, the maximum value of the proposed distance measure is the rank of the jNMF approximation, $k$.
However, $d(X_1, X_2)$ is frequently significantly below this value.
As a benchmark for considering the significance of these distance values, we measure the distance between our structured Swimmer image matrix $X_1$ and noise matrix $N$ to be $d(X_1, N) = 2.297$.
While this value is larger than other distance values observed in our previous experiments, is it far below the upper bound of $k=10$. 
Despite this relatively low distance measure, $d(X_1, N)$ can serve as a baseline for interpreting $d(X_1, X_2)$ for other matrices $X_2$.

We measure the distance between $X_1$ and a matrix $X_2$ formed by swapping the zeros and ones in $X_1$; see Figure~\ref{swimmer}.
Note that the data points in $X_1$ can be constructed by starting with the body and adding in limbs, while those in $X_2$ can be constructed by starting with a body with all possible limbs and covering the limbs that are not being used in a particular data point.
Figure~\ref{XandY} shows that this data set can be represented well with eight common basis vectors and two additional basis vectors, each of which is strongly associated with one data set.
The common basis vectors are used differently by the two data sets; in $X_1$, they are used to add limbs to the body, while in $X_2$, they are used to cover up limbs.
The method both identifies similar structures between the two datasets and extracts the features necessary distinguish them.

\begin{figure}[htp]
\centering
\begin{subfigure}[t]{0.1\textwidth}
    \centering
    \includegraphics[scale=0.1, angle=90]{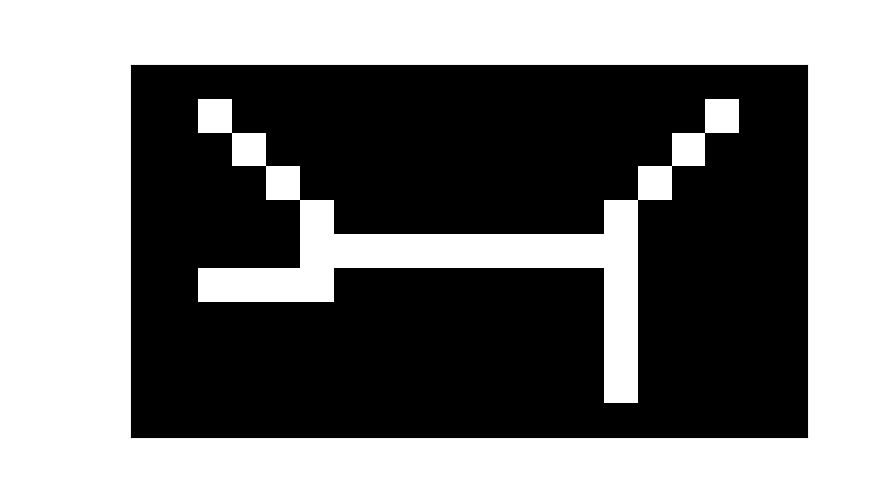}\\
    \includegraphics[scale=0.1, angle=90]{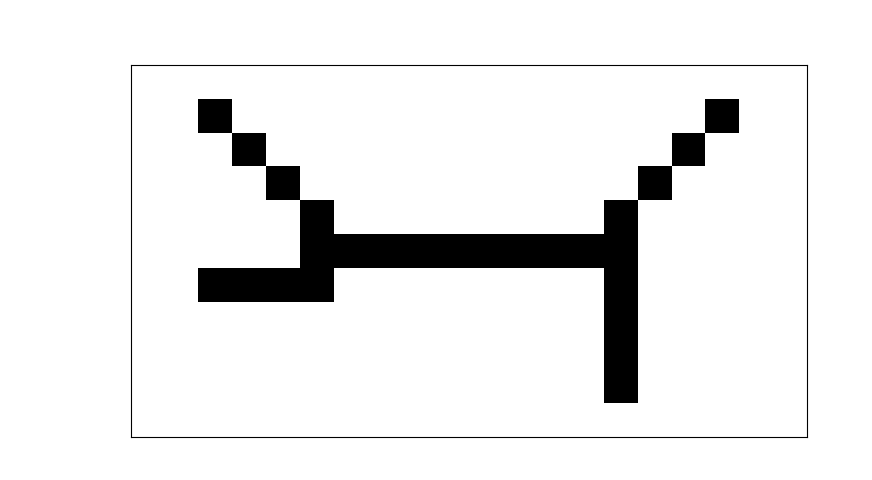}
    \caption{}\label{swimmer}
\end{subfigure}
\hfill
\begin{subfigure}[t]{0.38\textwidth}
    \centering
    \includegraphics[scale=0.1, angle=90]{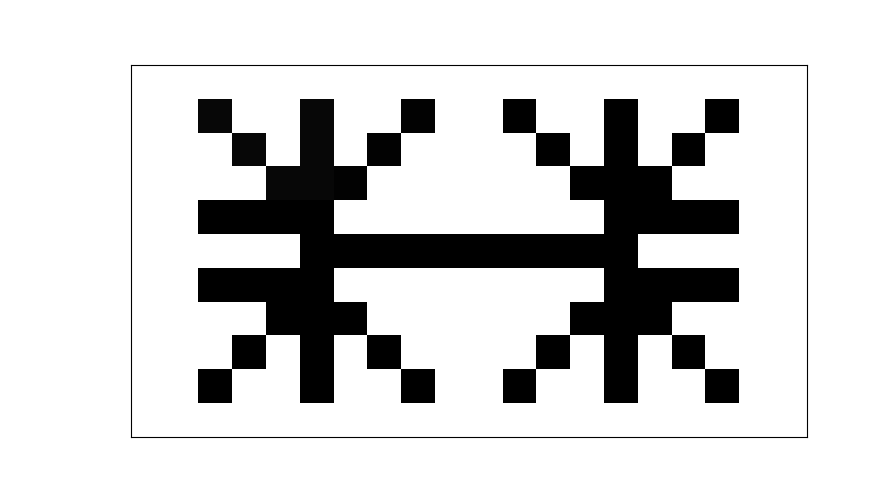}
    \includegraphics[scale=0.1, angle=90]{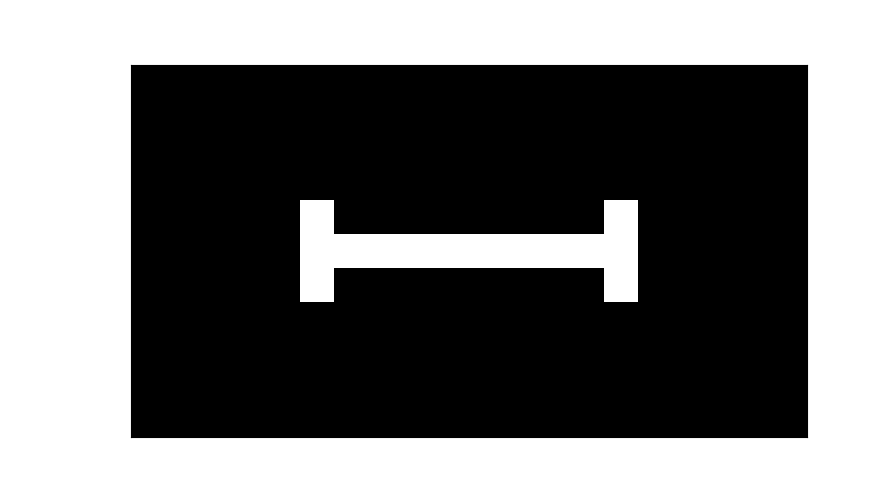}
    \includegraphics[scale=0.1, angle=90]{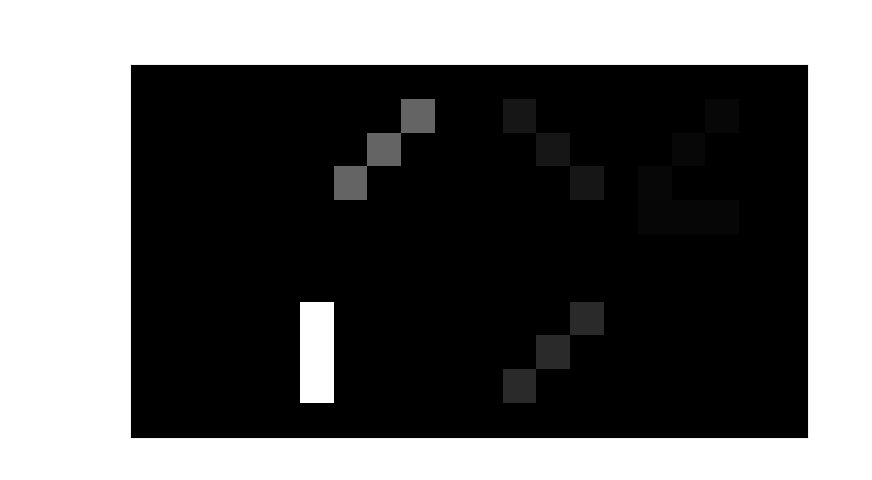}
    \includegraphics[scale=0.1, angle=90]{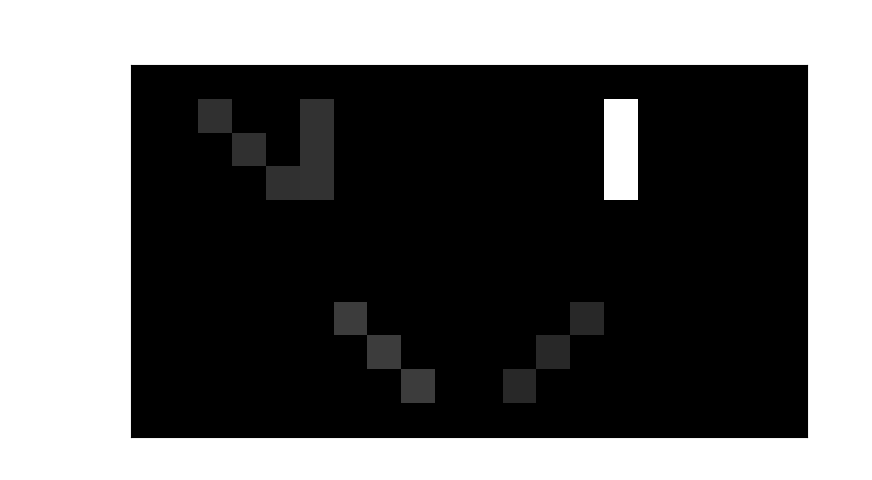}
    \includegraphics[scale=0.1, angle=90]{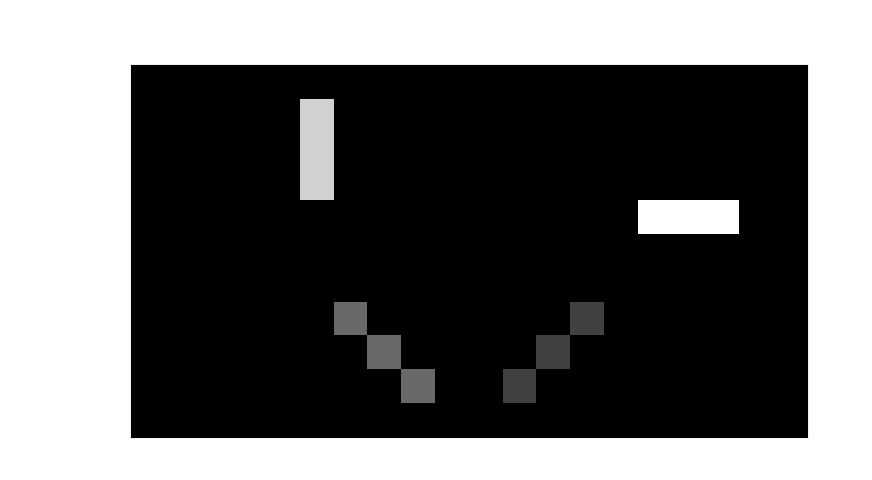}\\
    \includegraphics[scale=0.1, angle=90]{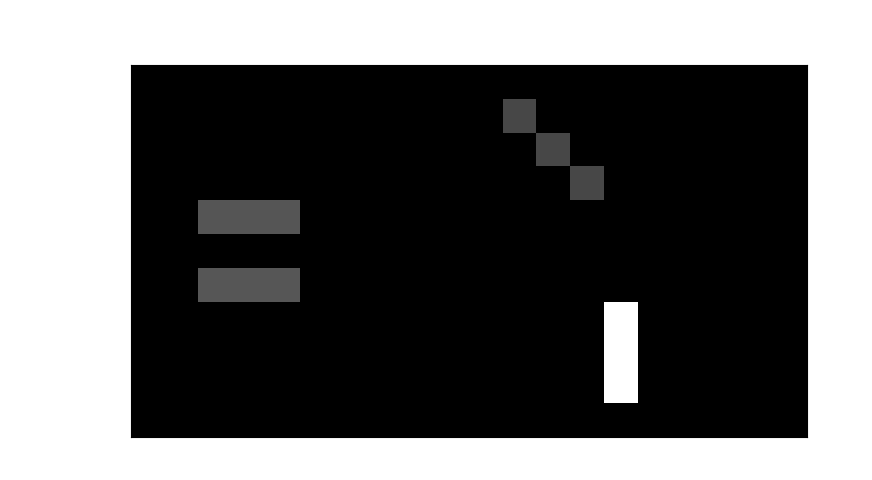}
    \includegraphics[scale=0.1, angle=90]{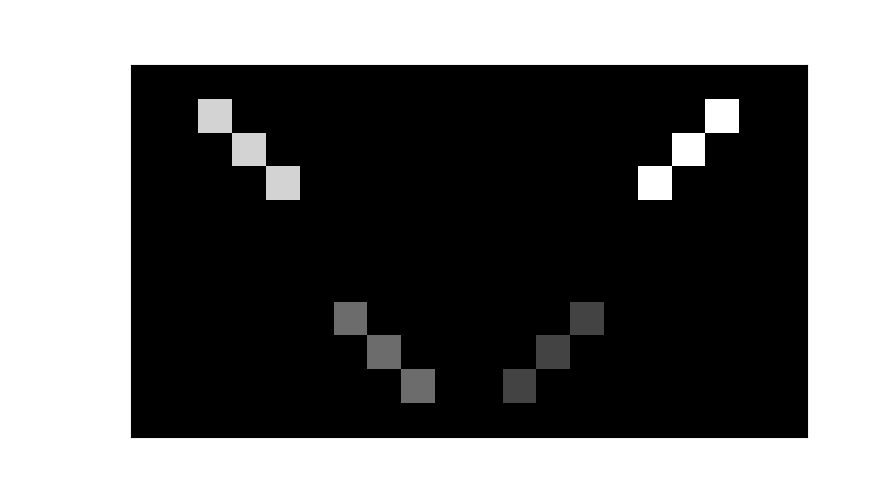}
    \includegraphics[scale=0.1, angle=90]{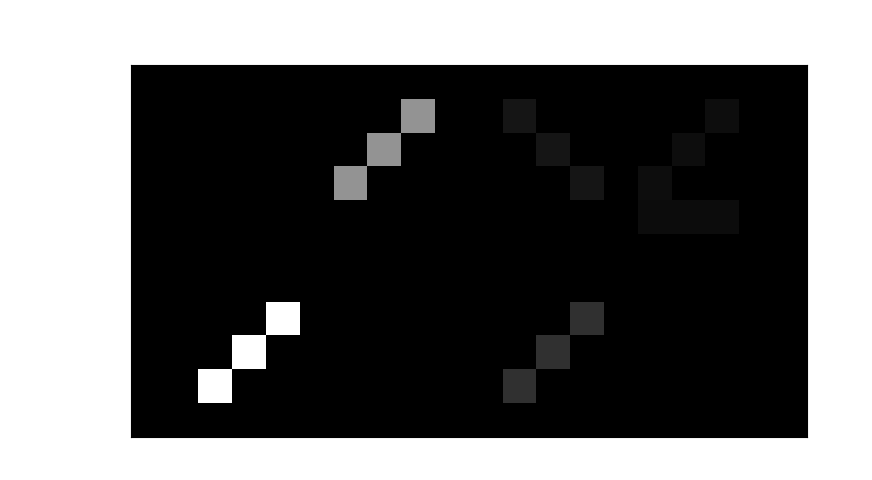}
    \includegraphics[scale=0.1, angle=90]{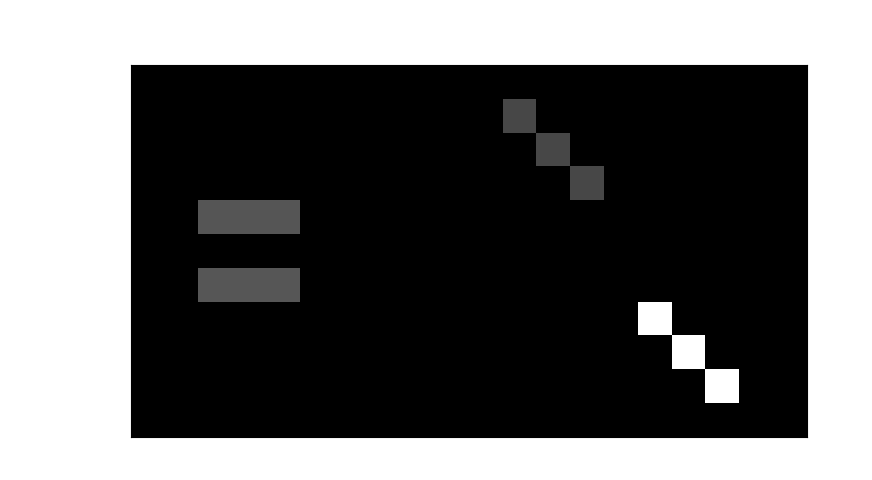}
    \includegraphics[scale=0.1, angle=90]{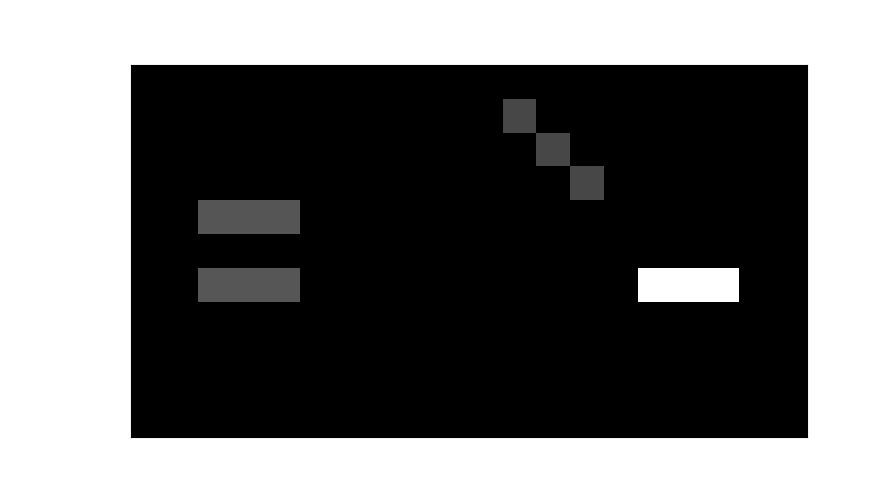}
    \caption{}
    \label{XandY}
\end{subfigure}
\caption{(a) Sample data points from the Swimmer data set (top) and the modified swimmer data set, where all zeros and ones are switched (bottom).
(b) Basis vectors learned by jNMF with rank $k=10$ on the Swimmer data set ($X_1$) with an inverted copy of the Swimmer data set such that all the zeros and ones are switched ($X_2$).
Ordering the basis vectors left to right and top to bottom, $\bar{\ve{p}}=[-0.999, 1.000, 0.010, -0.017, 0.003, -0.004, 0.015,$ $ 0.004, -0.001, -0.000]$ and $d(X_1, X_2) = 2.054$.
The leftmost basis vector in the top row contributes almost exclusively to $X_2$ and the basis vector in the second position of the top row contributes almost exclusively $X_1$. The other basis vectors contribute roughly evenly to both, adding limbs to $X_1$ and removing them from $X_2$.}
\end{figure}

\subsection{20 Newsgroups experiments}

As a final illustration of the promise of our proposed method and distance measure, we measure distances between the term frequency-inverse document frequency (tf-idf) representations of the various newsgroups (categories) in the 20 Newsgroups dataset~\cite{20news}.  The 20 Newsgroups dataset is a collection of approximately 20,000 newsgroup documents. The data set consists of six groups partitioned roughly according to subjects, with a total of 20 subgroups, and is an experimental benchmark for document classification and clustering; see e.g.,~\cite{lee2009semi}.

In Figures~\ref{fig:20_news} and~\ref{fig:20_news_chamfer}, we present heatmaps with colors corresponding to average jNMF distances and average Chamfer distances, respectively, between samples of 100 documents of each of the twenty newsgroups, averaged over 50 trials.  We remove headers, footers, and quotes from the 20 Newsgroups dataset, and then apply the tf-idf transformation to the entire set. In each trial, we sample 100 documents (represented as tf-idf vectors with length equal to the size of the entire data corpus) uniformly from each newsgroup and calculate pairwise distances between each sample. The rows and columns of the resulting distance matrix are then re-ordered and line-segregated according to cluster labels assigned by $k$-means with $k=6$ applied to the columns of the distance matrix.

Applying clustering to the jNMF and Chamfer distance matrices reveals existing block structure.  
While neither distance clustering respects the newsgroups divisions, the identified clusters represent highly related topics.
The clustering applied to the Chamfer distance matrix correctly identifies ``comp" newsgroup, while the jNMF clustering adds the ``for sale" group to this cluster.  
Both distance clusterings group the ``hockey" and ``baseball" groups.  Each clustering has an ``atheism"/``politics" cluster, but the jNMF clustering separates these into two clusters and includes the ``sci.med" group, while Chamfer groups ``sci.med" with ``sci.space" but places ``sci.crypt" into the ``atheism"/``politics" cluster.

Qualitatively, the clusters identified by the jNMF distance and the Chamfer distance are coherent.  However, the jNMF distance matrix produces clusters with significantly lower relative intra-cluster to inter-cluster distance ratio than that of the Chamfer distance.

\begin{figure}
    \centering
    \includegraphics[width=0.87\columnwidth]{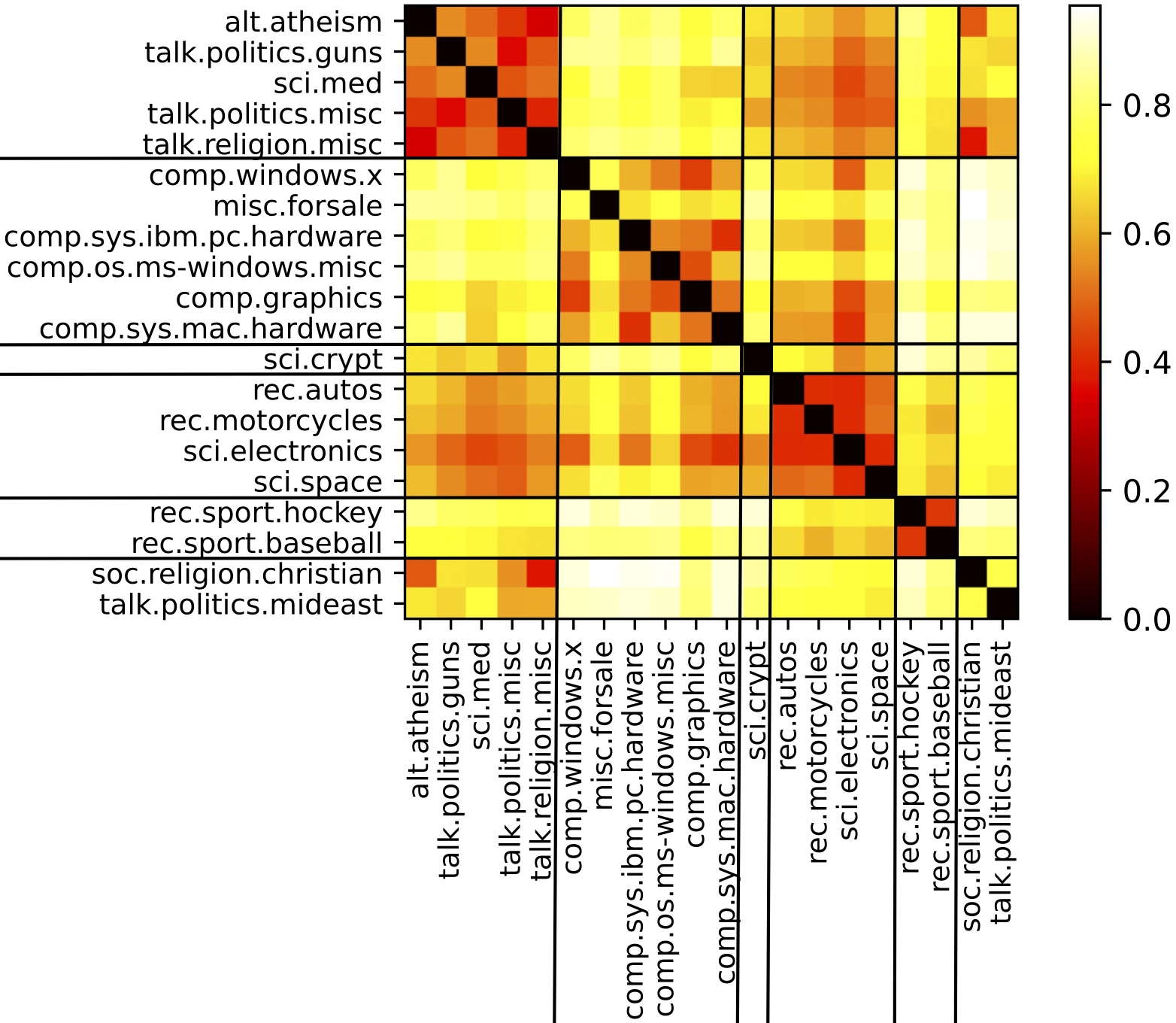}
    \caption{Average jNMF based distance between samples of 100 documents of the twenty newsgroups (averaged over 50 trials).}
    \label{fig:20_news}
\end{figure}

\begin{figure}
    \centering
    \includegraphics[width=0.9\columnwidth]{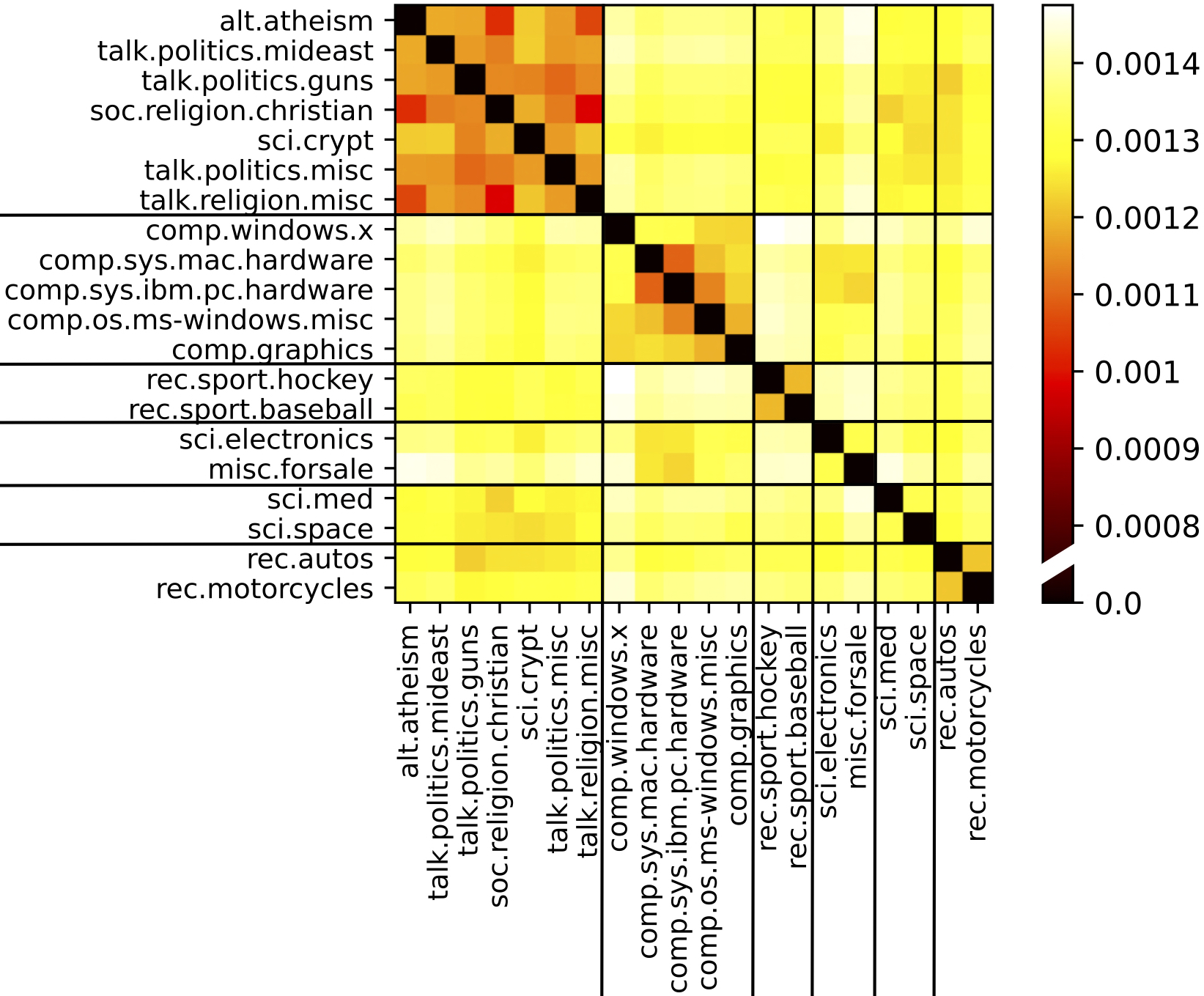}
    \caption{Average Chamfer distance between samples of 100 documents of the twenty newsgroups (averaged over 50 trials).}
    \label{fig:20_news_chamfer}
\end{figure}

\section*{Conclusion}
In this work, we proposed a promising distance measure for datasets based on shared features learned by jNMF. Our proposed distance measure indicates similarity of two datasets and the proposed method learns which basis components are shared between the datasets and which are not.
As one would hope, our proposed distance measure exhibits permutation and scaling invariance, symmetry, and monotonicity over subset relationships and additive noise in the data.

Future work includes applying the measure in tasks like anomaly or plagiarism detection, investigating hyperparameter choice, and exploring distance measures derived from different matrix factorizations or low-dimensional approximations.

\section*{Acknowledgments}
We thank April Zhao for helpful discussions early on.

\bibliographystyle{IEEEtran}
\bibliography{references}

\end{document}